\pdfoutput=1

\documentclass[11pt]{article}

\usepackage[preprint]{acl}

\usepackage{times}
\usepackage{latexsym}

\usepackage[T1]{fontenc}

\usepackage[utf8]{inputenc}

\usepackage{microtype}

\usepackage{inconsolata}

\usepackage{graphicx}
\usepackage{amssymb}
\usepackage{amsmath}
\usepackage{caption}
\usepackage{booktabs}
\usepackage{multirow}
\usepackage{subcaption}
\usepackage{makecell}
\usepackage{CJKutf8}

\title{\protect\includegraphics[height=0.45cm,width=0.4cm, trim=0cm 0.3cm 0cm 0cm, clip]{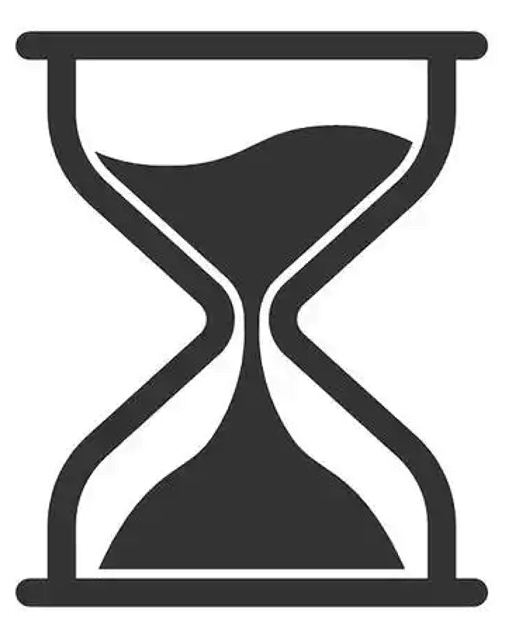} Temporal Alignment of LLMs through Cycle Encoding for Long-Range Time Representations}

\author{\small{Xue Han, Qian Hu, Yitong Wang, Wenchun Gao, Lianlian Zhang, Qing Wang, Lijun Mei, Chao Deng, Junlan Feng} \\
  JIUTIAN Team \\
  China Mobile Research Institute \\
  Beijing, China \\
  \texttt{\scriptsize{\{hanxueai, huqianai, wangyitongyjy, gaowenchun, zhanglianlian, wangqingai, meilijun, dengchao, fengjunlan\}@chinamobile.com}} \\}

\begin{document}
\maketitle
\begin{abstract}
Large language models (LLMs) suffer from temporal misalignment issues especially across long span of time. The issue stems from knowing that LLMs are trained on vast amounts of data with sparse temporal information over long periods, such as thousands of years, resulting in insufficient learning or catastrophic forgetting by the LLMs. This paper proposes a methodology named "Ticktack" for addressing the LLM's long-time span misalignment in a yearly setting. Specifically, we first propose to utilize the sexagenary year expression instead of the Gregorian year expression employed by LLMs, achieving a more uniform distribution in yearly granularity. Then, we employ polar coordinates to model the sexagenary cycle of 60 terms and the year order within each term, with additional temporal encoding to ensure LLMs understand them. Finally, we present a temporal representational alignment approach for post-training LLMs that effectively distinguishes time points with relevant knowledge, hence improving performance on time-related tasks, particularly over a long period. We also create a long time span benchmark for evaluation. Experimental results prove the effectiveness of our proposal.



\end{abstract}

\section{Introduction}

Language models have always suffered from temporal misalignment issues stemming from the temporal discrepancies between the training and testing data, resulting in variability in reference time during downstream tasks \citep{deepmind_2021,luu2022time,jaidka2018diachronic,tan2023towards}. The issues persist with recently released large language models (LLMs) such as LLama \citep{llama} and GPT-4 \citep{gpt4}, which are trained on enormous datasets and exhibit significant performance decreases over time, especially when the time periods are long \citep{set_the_clock,time_encoded_weight,luu2022time}.

\begin{figure}[ht]
\centering
\includegraphics[height=4.9cm,width=\linewidth,trim=0cm 0cm 1cm 1.5cm, clip]{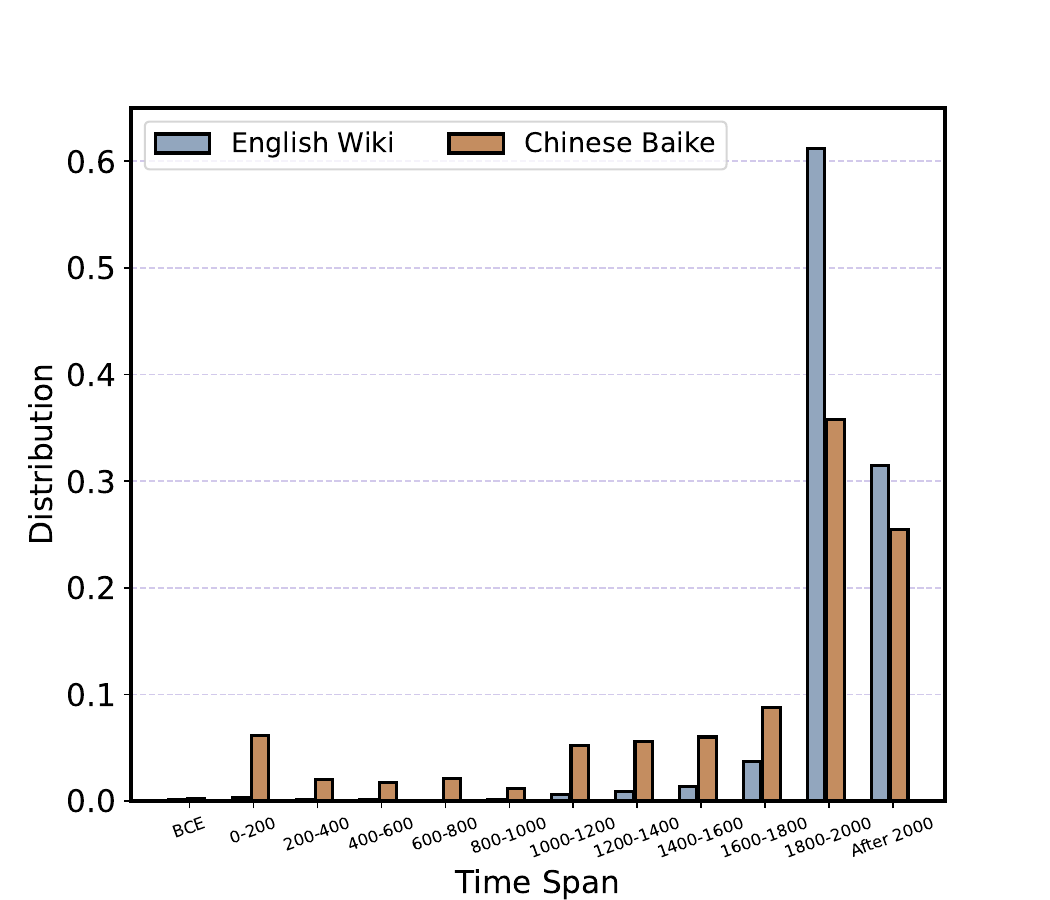}
\caption{The distribution of temporal information in both Wikipedia (English) and Baidu Baike (Chinese), with statistics conducted at intervals of 200 years from BCE to after 2000.}
\label{img_distribution}
\end{figure}

The long-span temporal misalignment issues in LLMs primarily arise from the extensive training data covering thousands of years (e.g., from BCE to post-2000 AD). The enormous training data generally lacks explicit temporal grounding, resulting in relatively limited and sparse knowledge of specific time periods \citep{time_gpt}. We investigate the distribution of years in the wiki dataset \footnote{\url{https://huggingface.co/datasets/wikimedia/wikipedia}} and Baidu Baike\footnote{\url{https://baike.baidu.com/}}, as illustrated in Figure \ref{img_distribution}. It indicates that data is rare in ancient ages, such as BCE, but concentrated in the internet era (1990s–present). Note that the year refers to the temporal reference of the data's content. Other studies \citep{once_time_graph} show similar findings. The sparse and long-tail distribution of training data over time results in insufficient learning or catastrophic forgetting in LLMs, leading to even poor performance during low-resource years \citep{mccoy2023embers,yasaman2022impact}.


Existing approaches \citep{editing1,editing2} to resolving time misalignment issues emphasize updating models with new knowledge, yet they do not assess the internal temporal knowledge of LLMs over long periods. The most relevant work for us is \citet{wei2025diachronic}, which divides the time span from Pre-Qin to Modern into three distinct periods. This classification is too coarse for reasoning in LLMs.

This paper proposes a plug-and-play methodology named "Ticktack" for addressing the LLM's long-time span misalignment in a yearly setting. \textbf{To begin}, we propose solving the sparse and long-tail distribution of training data throughout time by employing a novel sexagenary year expression instead of the Gregorian year expression used by LLMs. The idea is founded on the observations of \citet{tan2023towards} that the Gregorian year expression employed by LLMs resulted in an excessively wide range for the year embedding within the representation space of LLMs. Sexagenary time expression could achieve a more uniform distribution in yearly granularity. Figure \ref{img_dataset} in Section \ref{sec:exp_data} provides a comprehensive analysis. \textbf{Subsequently}, we apply polar coordinates to represent the sexagenary cycle of 60 terms and the chronological sequence inside each term, integrating additional temporal encoding to facilitate comprehension by LLMs. \textbf{Finally}, we present a temporal representational alignment approach for post-training LLMs that effectively distinguishes time points with relevant knowledge, thereby improving performance on time-related tasks, particularly over a long period.

Due to the lack of long time span benchmarks, we develop TempLS, a question-answering dataset covering the period from 75,000 BCE to 2025 AD, to facilitate the analysis of Ticktack's efficiency. We conduct experiments over several representative open-source LLMs ranging from 3 billion to 13 billion parameters. Experimental results on both open-source time-related benchmarks and TempLS prove the effectiveness of our proposal.

\section{Related work}
\textbf{Temporal expressions and embeddings in language models.} Traditional works \citep{once_time_graph} use a normalized value for time expression using tools such as SUTIME \citep{sutime}. With the development of pre-trained language models, researchers try to explore better time expression. In order to have a more comprehensive understanding of temporal expressions, \citet{tan2023towards} divifgfdes 1900 to 2040 into seven 20-year time periods. \citet{zhang2023mitigating} leverages duration statistics on each dataset’s development, such as seconds and minutes. However, all these time expressions within the scope of the Gregorian calendar system still suffer from the complicated representation space. \citet{wei2025diachronic} segments the Chinese lexicon history into three periods: Ancient, Middle Ancient, and Near Ancient, and uses a one-hot embedding to represent them. This classification is too coarse for reasoning in LLMs. Recent LLMs, such as GPT-4 \citep{gpt4}, tokenize numeric information independently from a perspective of tokenization. However, this approach lost the distinct meaning of terms like "2014" as a specific year. 

\textbf{Temporal alignment of language models.} Early efforts \citep{chew,templama} focus on designing novel datasets to probe LMs for temporal-related understanding, while more recent works \citep{tram, jang2022temporalwiki, benchdynamic} introduce new benchmarks to evaluate the temporal alignment capabilities. To create temporally adapted language models, conventional methods rely on continual learning on time-specific data \citep{agarwal2022temporal, loureiro2022timelms}. Lately, some knowledge modification techniques are proposed to align temporal knowledge \citep{zhu2020modifying, de2021editing, dai2021knowledge}. Other works \citep{set_the_clock, longpre2024pretrainer, gurnee} study the LMs’ temporal misalignment caused by the chaotic pretraining corpus \citep{longpre2024pretrainer} and LMs can represent temporal knowledge learned from pretraining in their internal states \citep{gurnee}. These findings open up the possibility of aligning models to a specific time.

\begin{figure}[ht]
\centering
\includegraphics[height=4.7cm,width=\linewidth]{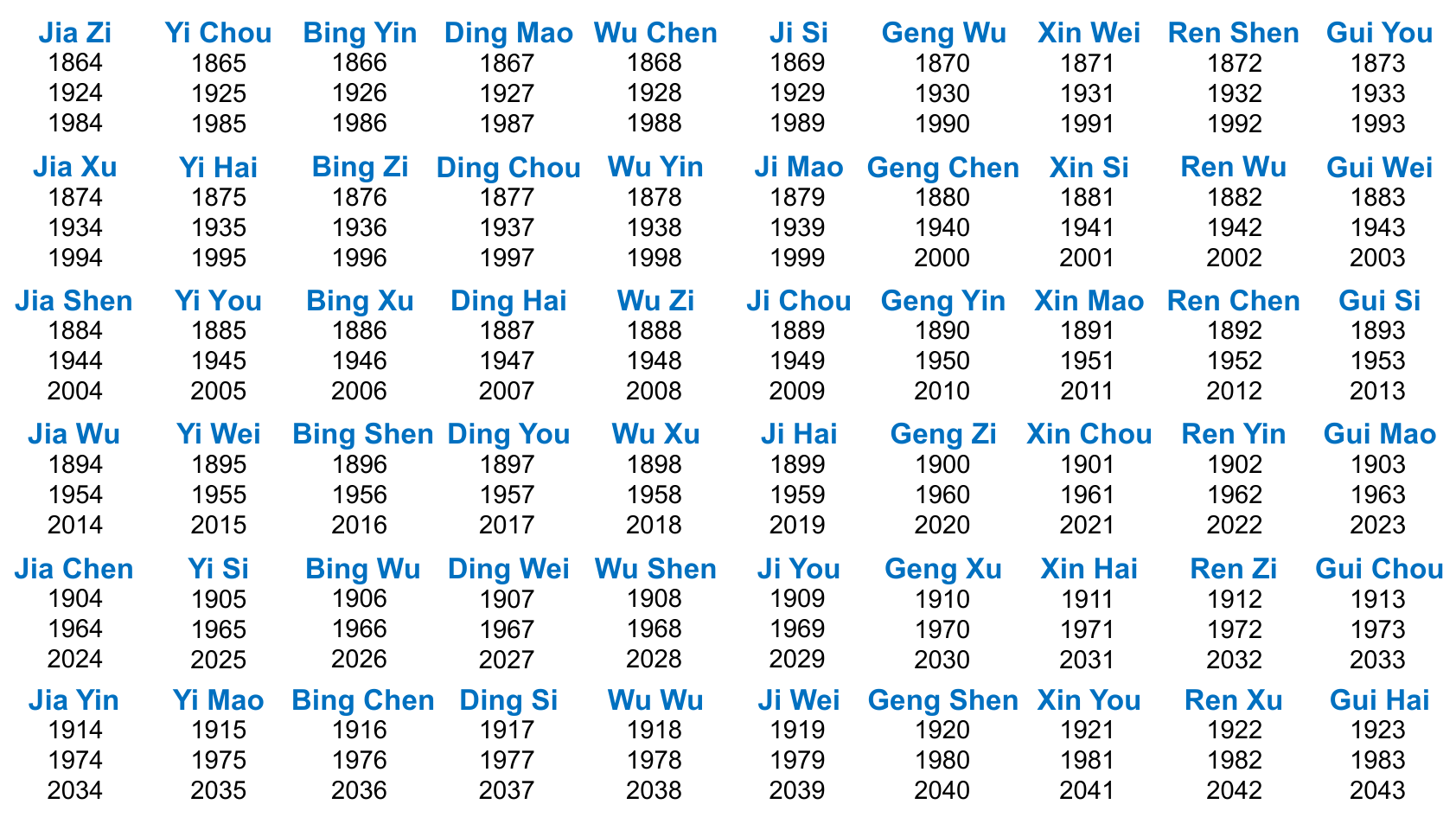}
\caption{The correspondence between the sexagenary year (blue) and Gregorian year (black). For instance, both 1864 and 1924 correspond to the "Jiazi" year.}
\label{img_example_tiangan}
\end{figure}


\begin{figure*}[ht]
\centering
\includegraphics[height=4.5cm,width=\linewidth]{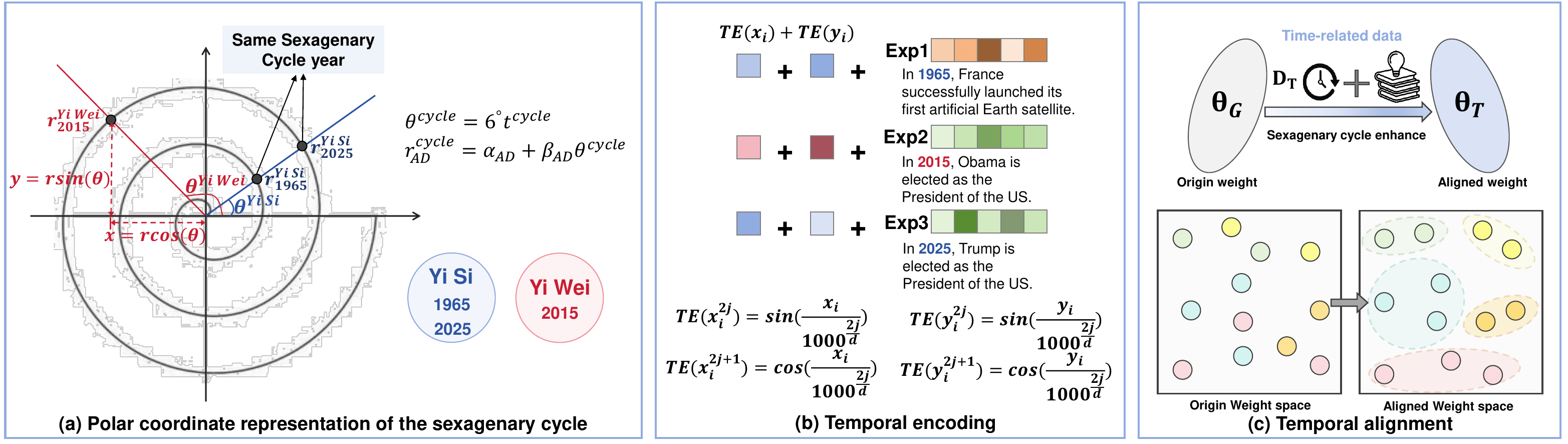}
\caption{An overview of Ticktack (a) illustrates the novel way to express the years, leveraging the polar coordinate representation of the sexagenary cycle. (b) adopts sine and cosine functions to encode temporal information based on the sexagenary cycle. (c) describes the temporal alignment process to further transform the original weight space of the LLMs into a temporal re-organized and distinguished weight space.}
\label{img_framework}
\end{figure*}

\section{Methodology}

\subsection{Preliminary: conversion between sexagenary year and Gregorian year}
We initially provide the concept of the sexagenary cycle chronology and its correlation with the Gregorian calendar year, which is primarily utilized by LLMs during data preprocessing. The sexagenary cycle generates a sixty-year cycle utilized in the calendars of China and several other Far Eastern nations, derived from the combination of two fundamental cycles of ten and twelve\footnote{\url{https://en.wikipedia.org/wiki/Sexagenary_cycle}}. 
Figure \ref{img_example_tiangan} depicts the sexagenary cycle chronology, which divides years into 60 categories, ranging from "Jiazi" to "Guihai." The associated Gregorian calendar years are grouped under their sexagenary category. For instance, the years 1864 AD and 1924 AD both correspond to the "Jiazi" year in the sexagenary cycle chronology.


By employing the sexagenary cycle chronology to represent the years, thousands of years of long-term data are reconstructed and aggregated into a 60-year cycle. As a result, the time representation achieves a more uniform distribution than the broader distribution space in the Gregorian year system, allowing for even better connections with relevant events. More detailed analyses are listed in Figure \ref{img_dataset} of Section \ref{sec:exp_data}.


\subsection{Overview of Ticktack}
Ticktack is a plug-and-play methodology for LLMs that translates and aligns the year representation from the Gregorian calendar system with the sexagenary cycle chronology, hence enhancing the performance of LLMs on temporal tasks over long spans. Figure \ref{img_framework} illustrates the pipelines of Ticktack. It consists of three modules: (a) a polar coordinate representation of the sexagenary years; (b) temporal encoding; and (c) temporal alignment of weight space.

Firstly, to make the LLMs understand the newly introduced sexagenary year expression, we utilize the polar coordinate to represent the sexagenary cycle of 60 terms and the Gregorian years order within each of the 60 categories. Then, we design a temporal encoding method to inject the newly introduced sexagenary year information into its associated input data embedding. Finally, we defined a temporal alignment objective to post-train the LLMs using the time-encoded input data. In this way, the pre-trained weight space of the LLMs is transformed into a temporal re-organized and distinguished weight space, enhancing the comprehension of long-span temporal information with relevant knowledge. In the next subsections, we will present the specifics for each module.


\subsection{Polar coordinate representation of the sexagenary year}
By using polar coordinates to represent the sexagenary year, we aim to trace the continuity of time information over a cycle period and bridge the representation similarity between years with a 60-year interval based on sexagenary cycle chronology. 

Given an input sequence $s_i$ of length $l$, the input embedding of $s_i$ generated by LLM is denoted as $h_i \in \mathbb{R}^{l\times d}$, where $d$ represents the hidden dimension. We define $t_i^{AD}$ to indicate the Gregorian year tokens in the input sequence $s_i$ (e.g. "$t_i^{AD}$ = 1965" in "France successfully launched its first artificial Earth satellite in 1965."). The Gregorian year $t_i^{AD}$ could be transformed to the sexagenary year $t_i^{cycle}$, according to the following Eq. \eqref{eq_cycle}:
\begin{equation}
t_i^{cycle}:\left\{
    \begin{aligned}
& (60 - |t_i^{AD}|-2) \ mod \ 60, \ t_i^{AD} <0\\
& (60 - |t_i^{AD}-3|) \ mod \ 60, \ 0<t_i^{AD}< 4 \\
& (t_i^{AD}-3) \ mod \ 60, \ t_i^{AD} \geq 4
\end{aligned}
\right.
\label{eq_cycle}
\end{equation}

To make the LLM understand the sexagenary year, we utilize the polar coordinate to represent $t_i^{cycle}$. As illustrated in Eq. \eqref{eq_polar}, we use $\theta_{cycle}$ to identify the 60 terms or categories in a sexagenary cycle. $r_{AD}^{cycle}$ is used to differentiate years within one category of a sexagenary cycle. That is to say, an ensemble of years within one category of a sexagenary cycle share the same angle but have different distances from the pole in polar coordinates. $\alpha_{AD}$ and $\beta_{AD}$ are hyperparameters to determine the $r_{AD}^{cycle}$.

\begin{equation}
\begin{split}
& r_{AD}^{cycle} = \alpha_{AD} +\beta_{AD} \theta_{cycle}\\
& \theta_{cycle} = \frac{360^{\circ}}{60}t_i^{cycle}=6^{\circ} t_i^{cycle}
\end{split}
\label{eq_polar}
\end{equation}

As seen in Figure \ref{img_framework}(a), 1965 AD and 2025 AD both belong to the "Yi Si" of the sexagenary year and hence share the same $\theta_{cycle}$ value. To distinguish them under "Yi Si," by adjusting the values of $\alpha_{AD}$ and $\beta_{AD}$, 2025 AD has a larger $r_{2025}^{Yi Si}$ compared to 1965 AD $r_{1965}^{Yi Si}$, making it farther away from the pole.

To better encode the temporal information later, we further convert the polar coordinate representation to the Cartesian coordinate system, as in Eq. (\ref{eq_cartesian}).

\begin{equation}
\begin{split}
& x_i = r_{AD}^{cycle}\cos(\theta_{cycle}), y_i = r_{AD}^{cycle}\sin(\theta_{cycle}) \\
\end{split}
\label{eq_cartesian}
\end{equation}
Where $x_i \in \mathbb{R}^d$ and $y_i \in \mathbb{R}^d$. 
$x_i$ and $y_i$ are the final representations used to encode the temporal information.

\subsection{Temporal encoding}

We adopt the sine and cosine functions to integrate the sexagenary year temporal information for the Transformer's \citep{tranformer} position encoding. To be specific, we define the temporal encoding $TE(x_i)$ and $TE(y_{i})$ for the $x_i$ and $y_i$ respectively, formulated as below:

\begin{equation}
\begin{split}
TE(x_{i}):\left\{
\begin{aligned}
& TE(x_{i}^{2j})=\sin (\frac{x_i}{10000^{\frac{2j}{d}}}) \\
& TE(x_{i}^{2j+1})=\cos (\frac{x_i}{10000^{\frac{2j}{d}}})
\end{aligned}
\right. \\
TE(y_{i}):\left\{
\begin{aligned}
& TE(y_{i}^{2j})=\sin (\frac{y_i}{10000^{\frac{2j}{d}}}) \\
& TE(y_{i}^{2j+1})=\cos (\frac{y_i}{10000^{\frac{2j}{d}}})
\end{aligned}
\right.
\end{split}
\label{eq_TE}
\end{equation}


The temporal encodings $TE(x_{i})$ and $TE(y_{i})$ are added to the original input embedding $h_i$, forming a new temporal enhanced input embedding $h_i^{'} \in \mathbb{R}^{l\times d}$ that includes sexagenary cycle chronological information, defined as below:
\begin{equation}
    h_i^{'} = h_i + TE(x_{i}) + TE(y_{i})
    \label{eq_concat}
\end{equation}

As shown in Exp2 and Exp3 in Figure \ref{img_framework}(b), despite their high similarity in the semantic representation of input embeddings, there exist distinct differences due to variations in time information. With the temporal encoding of sexagenary year, it is more prone to capturing this difference.

\subsection{Temporal alignment of LLMs' weight space}


Using the sexagenary cycle chronology to represent temporal information encoded in the input hidden embeddings, we propose a temporal alignment objective to further post-train the LLMs, transforming the LLMs' pre-trained weight space to a temporally enhanced new one and allowing the LLMs' representation to establish a linkage between learned knowledge and the related time period.


Given a set of $n$ hidden embeddings $H=\{h_1^{'},h_2^{'},...,h_n^{'}\}$ generated from the $n$ input sequences, each hidden embedding $h_i^{'}$ is constructed using the temporal encoding module's Eq. \eqref{eq_concat}. Through post-training on time-related texts, we aim to further transform the existing weight space $\theta_{G}$ of the trained LLM $M$ with general tasks into a time alignment weight space $\theta_{T}$, thereby enhancing the connection between the temporal information and learned knowledge. The definition of the transformation between weight spaces is as follows:
\begin{equation}
    L_{temporal}: M(h_i^{'};\theta^{G}) \to M(h_i^{t};\theta^{T})
    \label{eq_mapping}
\end{equation}

Where $h_i^{t} \in H^t=\{h_1^t,h_2^t,...,h_n^t\}$ represents the temporally aligned embeddings. $\theta^{G}$ and $\theta^{T}$ are the LLMs' pre-trained weight space and the one transformed after the temporal alignment objective function $L_{temporal}$.


To define the temporal alignment function $L_{temporal}$, we apply Elastic Weight Consolidation (EWC) theory \citep{ewc}, which is proposed to find a solution to a new task in the vicinity of an older one. In our scenario, EWC protects the general capabilities of the LLMs (simplified as Task $G$) by constraining the parameters $\theta^{T}$ of time-related tasks (simplified as Task $T$) utilizing a quadratic penalty to stay in a region of low error for the prior general task $G$ centered around its parameters $\theta^{G}$. According to EWC's theory, $L_{temporal}$ is defined as below:
\begin{equation}
 \mathcal{L}_{temporal}= \mathcal{L}_T(\theta^T)+\lambda \mathcal{L}[(\theta^G) \to (\theta^T)]
    \label{eq_ewc}
\end{equation}

$\mathcal{L}_T(\theta^T)$ is the loss for the task $T$ only. Task $T$ necessitates enhancement in the LLMs by transforming the weight space $\theta^G \to \theta^T$, while preserving the prior general knowledge of the LLMs. $\lambda$ sets how important the old task is compared to the new one.

To improve the LLMs' links between already memorized general knowledge and encoded sexagenary cycle temporal information, task $T$ uses the similarity algorithm to reassemble the hidden weight space based on the 60 categories in a sexagenary cycle. The setup is founded on \citet{time_encoded_weight}'s findings that "years or months that are closer together in time yield their embeddings that are also closer together in weight space." 

Specifically, there are a total of $\{1,2..,K\}(K=60)$ sexagenary year classes, and each $h_i^t$ is assigned to one of these class $k$ based on $t_i^{cycle}$. For the embeddings $H_t^k $ in the class $k$, $H_t^k=\{h_1^t, h_2^t,...,h_m^t\}(m<n, H_t^{k} \subseteq H^t)$, our goal is to minimize the distance between embeddings in the intra-class while maximizing the separation between embeddings in the inter-classes, thus the objective of task $T$ is defined as:
\begin{equation}
\begin{split}
\mathcal{L}_{T} &= \delta \mathcal{L}_{intra} + (1-\delta)\mathcal{L}_{inter} \\
\mathcal{L}_{intra} &= 1-\sum \limits_{h_i^t \in H_t^{k}} \sum \limits_{h_j^t \in H_t^{k}} cos\_sim(h_i^t,h_j^t) \\
\mathcal{L}_{inter} &= \sum \limits_{h_i^t \in H_t^{k}} \sum \limits_{h_j^t \notin H_t^{k}} cos\_sim(h_i^t,h_j^t) \\
\end{split}
    \label{eq_loss_sim}
\end{equation}

Based on Eq. \eqref{eq_ewc}, the target for transformation between weight spaces is to minimize the objective below: 
\begin{equation}
\mathcal{L}_{temporal} = \mathcal{L}_{T} +
\frac{\lambda}{2}\mathcal{F}(\theta^T-\theta^G)^2
    \label{eq_loss}
\end{equation}

Where $\mathcal{F}$ is considered as the Fisher information matrix \citep{fisher1922mathematical}, for it is thus easy to calculate even for large models.

The ultimate loss ${L}_{final}$ comprises the generating objective of the LLMs (predicting the next token) ${L}_{NTP}$ and the temporal alignment objective ${L}_{temporal}$, defined as below:
\begin{equation}
\mathcal{L}_{final} = \mathcal{L}_{NTP} + \sigma \mathcal{L}_{temporal}
    \label{eq_loss_final}
\end{equation}

Where $\sigma$ controls the influence of the sexagenary cycle.

\begin{figure}[ht]
	\centering
	\begin{minipage}{\linewidth}
		\centering
		\includegraphics[height=5.2cm,width=0.9\linewidth,trim=1.5cm 1cm 1.2cm 0.5cm, clip]{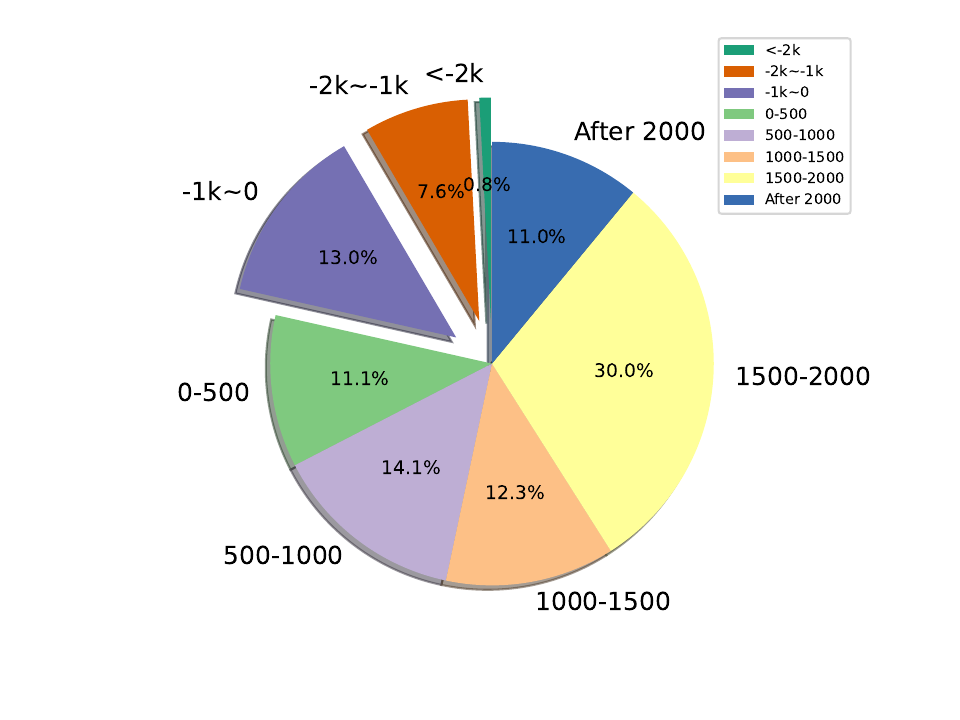}
		\label{dataset1}
	\end{minipage}
	
	\begin{minipage}{\linewidth}
		\centering
	\includegraphics[height=4.7cm,width=0.7\linewidth,trim=5.0cm 4.7cm 3.8cm 5.1cm, clip]{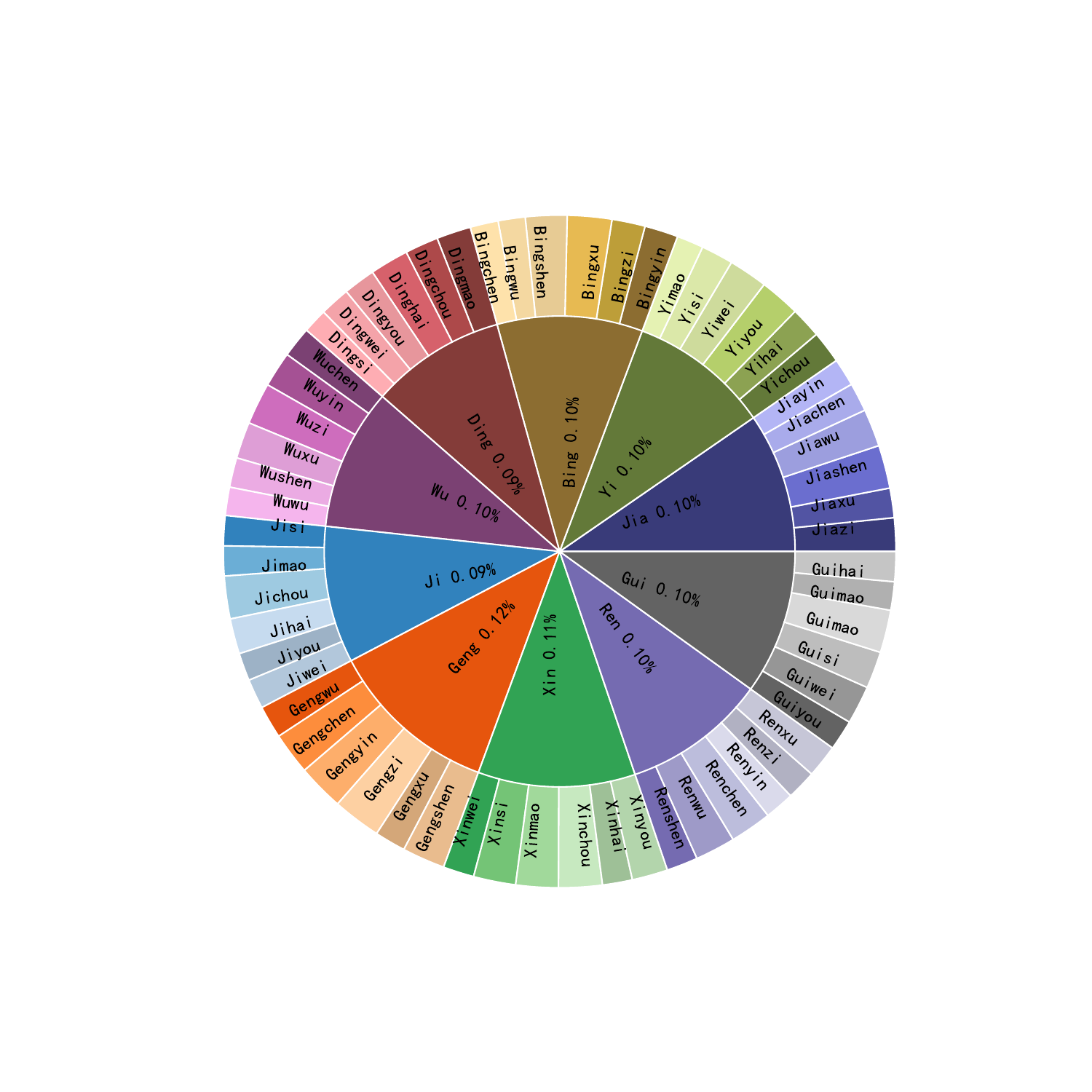}
		\label{dataset2}
	\end{minipage}
 \caption{The distribution of the years in our constructed TempLS dataset. The above figure summarizes the distribution of Gregorian years. The figure below displays the distribution of sexagenary years, which is apparently more uniform.} 
 \label{img_dataset}
\end{figure}

\begin{table*}[ht]
\centering
\caption{\label{table_main} Zero-shot and few-shot (5-shot) results of LLMs measured on TempLAMA, TempUN, and TempLS. Best performance is marked as bold. \textbf{(w/ PT)}: post-train base model with the predict-next prediction objective. \textbf{(w/ Ticktack)} is the temporal enhanced model with our proposal.}
\setlength{\tabcolsep}{1.5mm}{
\begin{tabular}{r|r|ll|llll|ll}
\hline
\hline
\multicolumn{2}{c}{\multirow{2}{*}{\textbf{Tasks}}}  & \multicolumn{2}{c}{\textbf{TempLS}} & \multicolumn{4}{c}{\textbf{TempLAMA}}  & \multicolumn{2}{c}{\textbf{TempUN}} \\ \cmidrule(r){3-4}  \cmidrule(r){5-8} \cmidrule(r){9-10}
\multicolumn{1}{l}{}   & \multicolumn{1}{l}{}               & \multicolumn{1}{l}{\small{\textbf{Zero-shot}}}                             & \multicolumn{1}{l}{\small{\textbf{Few-shot}}}                              & \multicolumn{2}{c}{\small{\textbf{Zero-shot}}}                             & \multicolumn{2}{c}{\small{\textbf{Few-shot}}}                              & \multicolumn{1}{l}{\small{\textbf{Zero-shot}}}                             & \multicolumn{1}{l}{\small{\textbf{Few-shot}}}                              \\ \hline
\multicolumn{2}{c|}{\textbf{Model}} & \multicolumn{1}{c}{\small{\textbf{Acc.}}} & \multicolumn{1}{c|}{\small{\textbf{Acc.}}}  & \multicolumn{1}{c}{\small{\textbf{ROUGE}}} & \multicolumn{1}{c}{\small{\textbf{F1}}} & \multicolumn{1}{c}{\small{\textbf{ROUGE}}} & \multicolumn{1}{c|}{\small{\textbf{F1}}} & \multicolumn{1}{c}{\small{\textbf{Acc.}}} & \multicolumn{1}{c}{\small{\textbf{Acc.}}} \\ \hline
\multirow{3}{*}{\textbf{\small{Qwen2.5-3B}}} & \small{\textbf{Base}}  & \multicolumn{1}{c}{62.37}   & \multicolumn{1}{c|}{67.81} & \multicolumn{1}{c}{17.13}  &  \multicolumn{1}{c}{7.45} & \multicolumn{1}{c}{17.14}  & \multicolumn{1}{c|}{7.45} & \multicolumn{1}{c}{59.25}   &  \multicolumn{1}{c}{58.88} \\
& \small{\textbf{w/ PT}}  & \multicolumn{1}{c}{67.30}   & \multicolumn{1}{c|}{\textbf{68.29}} & \multicolumn{1}{c}{53.76}  &  \multicolumn{1}{c}{33.36} & \multicolumn{1}{c}{53.73}  & \multicolumn{1}{c|}{33.43} & \multicolumn{1}{c}{44.82}   &  \multicolumn{1}{c}{28.05}   \\
& \small{\textbf{w/ Ticktack}}  & \multicolumn{1}{c}{\textbf{67.63}}   & \multicolumn{1}{c|}{67.62} & \multicolumn{1}{c}{\textbf{54.65}}  &  \multicolumn{1}{c}{\textbf{36.43}} & \multicolumn{1}{c}{\textbf{54.66}}  & \multicolumn{1}{c|}{\textbf{36.44}} & \multicolumn{1}{c}{\textbf{64.96}}   &  \multicolumn{1}{c}{\textbf{59.92}}   \\ \hline
\multirow{3}{*}{\textbf{\small{Qwen2.5-7B}}} & \small{\textbf{Base}}  & \multicolumn{1}{c}{73.66}   & \multicolumn{1}{c|}{72.89} & \multicolumn{1}{c}{14.12}  &  \multicolumn{1}{c}{6.66} & \multicolumn{1}{c}{14.12}  & \multicolumn{1}{c|}{6.66} & \multicolumn{1}{c}{57.20}   &  \multicolumn{1}{c}{\textbf{75.60}} \\
& \small{\textbf{w/ PT}}  & \multicolumn{1}{c}{78.12}   & \multicolumn{1}{c|}{78.17} & \multicolumn{1}{c}{50.97}  &  \multicolumn{1}{c}{27.19} & \multicolumn{1}{c}{50.97}  & \multicolumn{1}{c|}{27.19} & \multicolumn{1}{c}{67.79}   &  \multicolumn{1}{c}{58.22}   \\
& \small{\textbf{w/ Ticktack}}  & \multicolumn{1}{c}{\textbf{82.82}}   & \multicolumn{1}{c|}{\textbf{83.29}} & \multicolumn{1}{c}{\textbf{54.82}}  &  \multicolumn{1}{c}{\textbf{27.41}} & \multicolumn{1}{c}{\textbf{54.82}}  & \multicolumn{1}{c|}{\textbf{28.85}} & \multicolumn{1}{c}{\textbf{74.29}}   &  \multicolumn{1}{c}{74.37}   \\ \hline
\multirow{3}{*}{\textbf{\small{LLaMA2-7B}}} & \small{\textbf{Base}}  & \multicolumn{1}{c}{21.54}   & \multicolumn{1}{c|}{48.52} & \multicolumn{1}{c}{10.14}  &  \multicolumn{1}{c}{4.50} & \multicolumn{1}{c}{10.14}  & \multicolumn{1}{c|}{4.50} & \multicolumn{1}{c}{15.65}   &  \multicolumn{1}{c}{13.31} \\
& \small{\textbf{w/ PT}}  & \multicolumn{1}{c}{37.61}   & \multicolumn{1}{c|}{51.22} & \multicolumn{1}{c}{37.09}  &  \multicolumn{1}{c}{18.44} & \multicolumn{1}{c}{37.09}  & \multicolumn{1}{c|}{18.44} & \multicolumn{1}{c}{16.33}   &  \multicolumn{1}{c}{11.84}   \\
& \small{\textbf{w/ Ticktack}}  & \multicolumn{1}{c}{\textbf{58.45}}   & \multicolumn{1}{c|}{\textbf{59.18}} & \multicolumn{1}{c}{\textbf{45.42}}  &  \multicolumn{1}{c}{\textbf{23.90}} & \multicolumn{1}{c}{\textbf{45.42}}  & \multicolumn{1}{c|}{\textbf{23.90}} & \multicolumn{1}{c}{\textbf{25.18}}   &  \multicolumn{1}{c}{\textbf{25.14}}   \\ \cline{1-2}
\multirow{2}{*}{\textbf{\small{\makecell[c]{TimeLLaMA-7B}}}} & \small{\textbf{Base}}  & \multicolumn{1}{c}{24.46}   & \multicolumn{1}{c|}{43.59} & \multicolumn{1}{c}{0.00}  &  \multicolumn{1}{c}{0.00} & \multicolumn{1}{c}{0.00}  & \multicolumn{1}{c|}{0.00} & \multicolumn{1}{c}{14.24}   &  \multicolumn{1}{c}{22.44} \\
& \small{\textbf{w/ PT}}  & \multicolumn{1}{c}{43.54}   & \multicolumn{1}{c|}{47.61} & \multicolumn{1}{c}{34.95}  &  \multicolumn{1}{c}{22.61} & \multicolumn{1}{c}{35.82}  & \multicolumn{1}{c|}{23.06} & \multicolumn{1}{c}{18.84}   &  \multicolumn{1}{c}{18.89}   \\ \hline
\multirow{3}{*}{\textbf{\small{LLaMA2-13B}}} & \small{\textbf{Base}}  & \multicolumn{1}{c}{32.63}   & \multicolumn{1}{c|}{33.97} & \multicolumn{1}{c}{13.99}  &  \multicolumn{1}{c}{6.22} & \multicolumn{1}{c}{13.99}  & \multicolumn{1}{c|}{6.22} & \multicolumn{1}{c}{20.88}   &  \multicolumn{1}{c}{\textbf{26.07}} \\
& \small{\textbf{w/ PT}}  & \multicolumn{1}{c}{24.08}   & \multicolumn{1}{c|}{43.21} & \multicolumn{1}{c}{55.55}  &  \multicolumn{1}{c}{27.67} & \multicolumn{1}{c}{55.55}  & \multicolumn{1}{c|}{27.67} & \multicolumn{1}{c}{12.73}   &  \multicolumn{1}{c}{22.40}   \\
& \small{\textbf{w/ Ticktack}}  & \multicolumn{1}{c}{\textbf{65.28}}   & \multicolumn{1}{c|}{\textbf{70.18}} & \multicolumn{1}{c}{\textbf{62.63}}  &  \multicolumn{1}{c}{\textbf{32.96}} & \multicolumn{1}{c}{\textbf{62.63}}  & \multicolumn{1}{c|}{\textbf{32.96}} & \multicolumn{1}{c}{\textbf{25.36}}   &  \multicolumn{1}{c}{{25.36}}   \\ \cline{1-2}
\multirow{2}{*}{\textbf{\small{\makecell[c]{TimeLLaMA-13B}}}} & \small{\textbf{Base}}  & \multicolumn{1}{c}{52.55}   & \multicolumn{1}{c|}{53.71} & \multicolumn{1}{c}{0.00}  &  \multicolumn{1}{c}{0.00} & \multicolumn{1}{c}{0.00}  & \multicolumn{1}{c|}{0.00} & \multicolumn{1}{c}{24.78}   &  \multicolumn{1}{c}{24.37} \\
& \small{\textbf{w/ PT}}  & \multicolumn{1}{c}{53.00}   & \multicolumn{1}{c|}{54.24} & \multicolumn{1}{c}{43.77}  &  \multicolumn{1}{c}{24.40} & \multicolumn{1}{c}{43.77}  & \multicolumn{1}{c|}{24.40} & \multicolumn{1}{c}{23.85}   &  \multicolumn{1}{c}{14.53}   \\ \hline
\hline
\end{tabular}}
\end{table*}

\section{Experiments}

\subsection{Datasets and downstream tasks}
\label{sec:exp_data}

To evaluate the LLMs’ ability to understand temporal information, we utilize two typical temporal question-answering (QA) downstream tasks: \textbf{TempLAMA} \citep{templama} and \textbf{TempUN} \citep{beniwal2024remember}. TempLAMA is a time-sensitive QA dataset constructed based on Wikidata. TempUN is a large temporal multiple-choice QA dataset constructed by curating temporal information from the "Our World in Data" website.

To evaluate Ticktack's performance on the long-span time challenge, we create \textbf{TempLS}, as existing benchmarks predominantly focus on the internet age, to the best of our knowledge. \textbf{TempLS} is a long-span Chinese time-related multiple-choice QA dataset, including 137,090 question-answer pairs extracted by following the below steps.
Firstly, time-related texts covering a time span from 75,000 BCE to 2025 AD are filtered from the Baidu Baike. Then the filtered texts are summarized and converted into a QA format using Qwen57B\footnote{\label{qwen_web}\url{https://github.com/QwenLM/Qwen2.5}}. The specific distribution with the years within the dataset is depicted in Figure \ref{img_dataset}. It is apparent that with sexagenary year representation, the QA pairs have a more uniform distribution.

Appendix \ref{datasets} provides detailed information and samples of the aforementioned datasets.

\subsection{Experimental setups}
\textbf{Baselines models.}
To demonstrate the generality of our method, we select several representative open-source LLMs as base models, including Qwen2.5-3B\footref{qwen_web}, Qwen2.5-7B\footref{qwen_web}\citep{qwen2}, LLaMA2-7B\footnote{\label{llama_web}\url{https://ai.meta.com/resources/models-and-libraries/llama/}},  LLaMA2-13B\footref{llama_web}\citep{llama2}.

We post-train the above models with our proposed Ticktack for the temporal alignments of base LLMs, denoted as \textbf{(w/ Ticktack)}, and then evaluate their performance on the three temporal QA tasks mentioned above. 
For TempLS, we split the post-training dataset into 132,830 training samples and 4,260 testing samples; TempLAMA and TempUN follow their existing splits.

To comparison, we post-train all base LLMs with the typical next token prediction objective, referred to as \textbf{(w/ PT)}, using the exact same post-training dataset as the Ticktack post-training. We also compare with the open-source LLM series \textbf{TimeLlama-7b} and \textbf{TimeLlama-13b}\citep{TimeLlaMA}, which are optimized for temporal reasoning utilizing LlaMA2.

\begin{figure}[ht]
	\centering
	\begin{minipage}{\linewidth}
		\centering
		\includegraphics[height=4.5cm,width=6.3cm, trim=0.8cm 0.8cm 3.cm 1.0cm, clip]{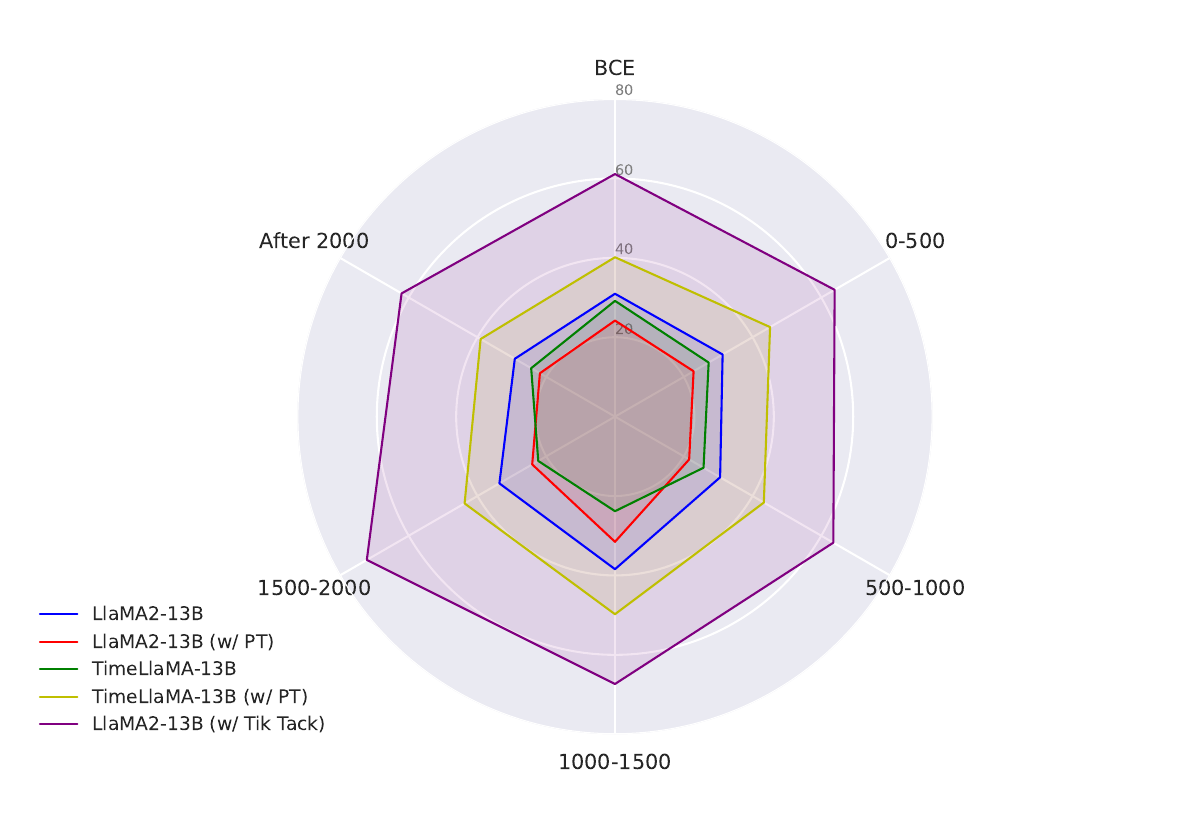}
		\label{img_radar_zero}
	\end{minipage}
	
	\begin{minipage}{\linewidth}
		\centering
	\includegraphics[height=4.5cm,width=6.3cm, trim=0.8cm 0.8cm 3.cm 1.0cm, clip]{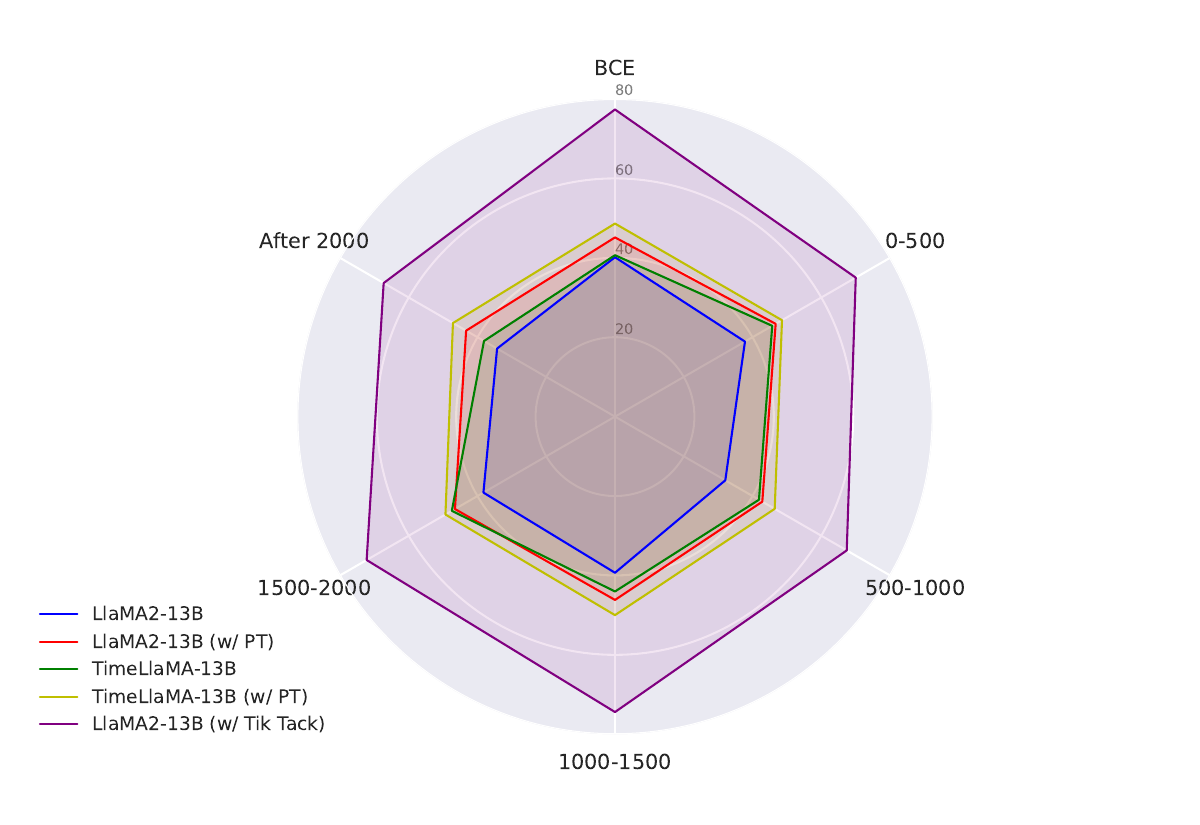}
		\label{img_radar_few}
	\end{minipage}
 \caption{Accuracy of Zero-Shot and Few-shot evaluations on the TempLS for the time-span from years BCE to after 2000.} 
 \label{img_radar}
\end{figure}

\textbf{Implementation details.} We employ the parameter expansion technique, known as LoRA \citep{lora}, for the post-training strategy of Ticktack. We are freezing the pre-trained parameters of the LLMs while incorporating trainable rank-decomposition matrices into each layer. $\alpha_{AD}$ and $\beta_{AD}$ in Eq. \eqref{eq_cartesian} are set to [0.5-1.0] and [0.5-1.0] for changes of them within this range similarly affect the value of the Cartesian coordinate according to our experiments. $\sigma$ in Eq. \eqref{eq_loss_sim} is set to 0.5. $\delta$ in Eq. \eqref{eq_loss_final} is affected by the batch size and set to 1 in current experiments. The hyperparameters employed for all base models are as follows: a batch size of 8, gradient accumulation steps of 2, 10 epochs, and a learning rate of $10^{-4}$. During the validation of downstream tasks, we utilize zero-shot and 5-shot settings to evaluate the best performance of the models.  Our setup consists of a four-core CPU and eight NVIDIA Tesla A100 GPUs.


\subsection{Results and analysis}
\textbf{Performance on downstream tasks.} Table \ref{table_main} presents the zero-shot and few-shot experimental results for the three temporal downstream tasks. Notably, each model trained using our proposed Ticktack significantly outperforms its baseline and post-trained counterparts across various performance metrics. In comparison to post-training alone (w/ PT), the model trained with Ticktack (w/ Ticktack) exhibits remarkable enhancements across nearly all measures in the three downstream tasks. Ticktack demonstrates an average accuracy increase of 34.43\% on TempLS in comparison with the base LLaMA2-13B. In contrast to the comparative temporal enhanced baselines TimeLLaMA-7B and TimeLLaMA-13B, our method for both 7B and 13B scales achieves better results on most evaluation metrics. Furthermore, due to its inadaptability with the format of the TempLAMA task (TempLAMA is the only task whose answer is not choice, detailed in Table \ref{tab:examples} of Appendix\ref{sec:appendix}), the evaluation results of TimeLLaMa models on this task are 0. After post-training, the performance of TimeLLaMa models could improve on TempLAMA. Particularly, Ticktack demonstrates a more significant enhancement on long-span datasets TempLS compared to most of the baseline models. 


\begin{figure}[ht]
\centering
\includegraphics[height=9.5cm,width=5.2cm]{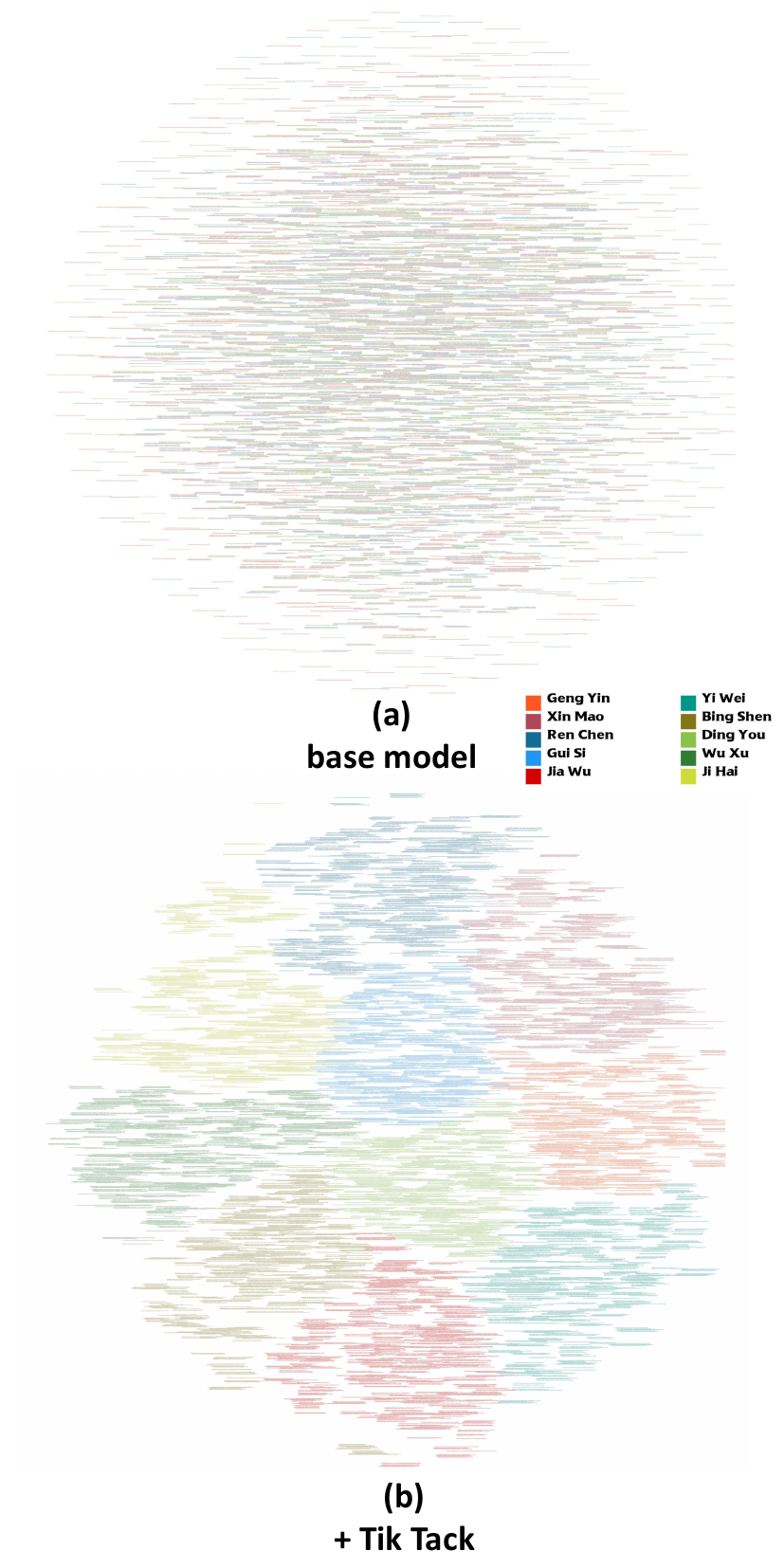}
\caption{The distribution of Qwen-3B's outputs of 10,000 sentence vectors. (a) The pre-trained embeddings are dispersed throughout the vector space, which is hard to distinguish. (b) After post-training by Ticktack, the temporal enhanced embeddings exhibit clustering characteristics according to years.}
\label{img_tsne}
\end{figure}

\begin{figure*}[ht]
  \centering 
  \subfloat[Base model] 
  {
      \includegraphics[height=4.3cm,width=5.4cm, trim=2.0cm 0.8cm 3.cm 1.0cm, clip]{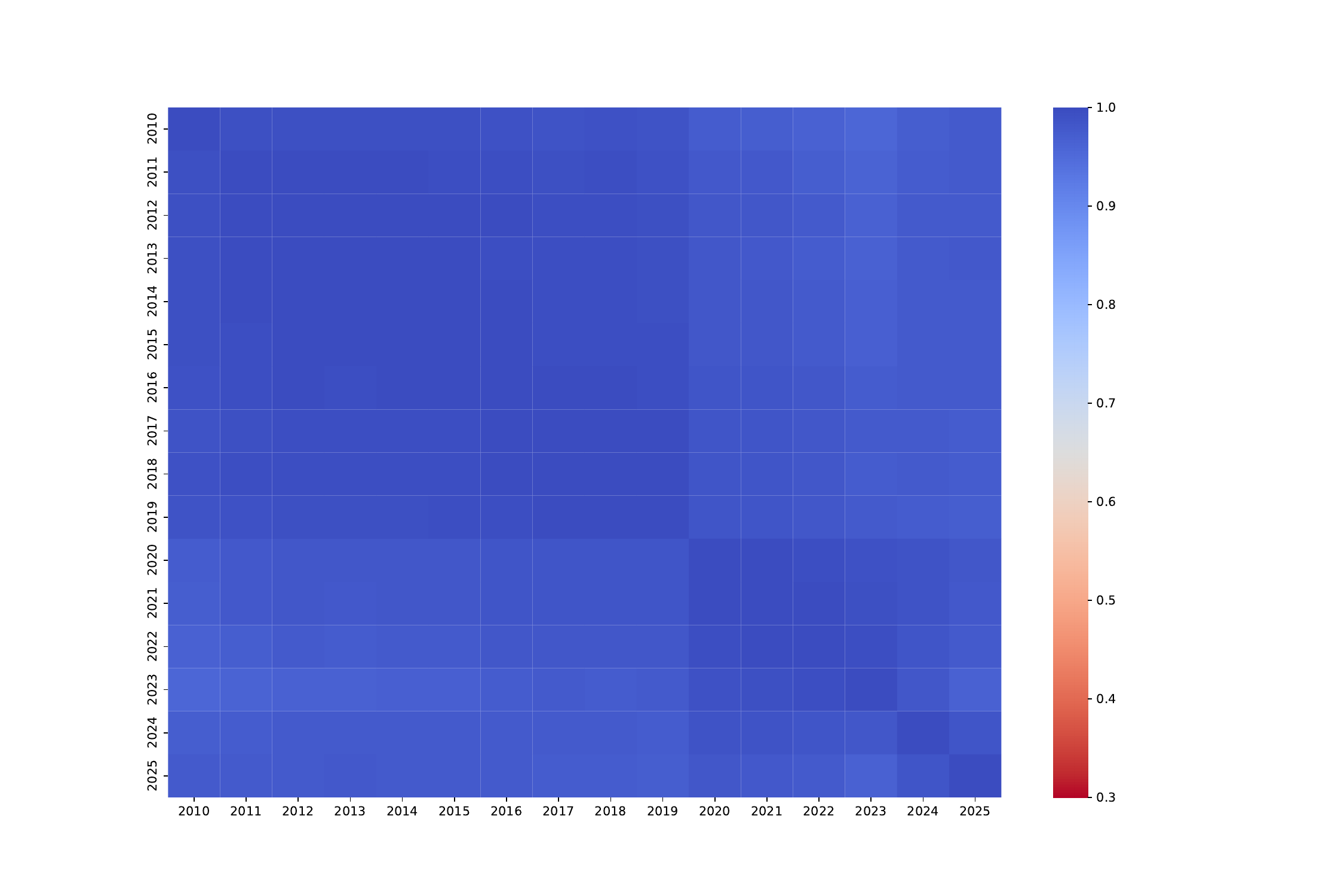}
  }
  \subfloat[Temporal encoding]
  {
      \includegraphics[height=4.3cm,width=5.4cm, trim=2.0cm 0.8cm 3.cm 1.0cm, clip]{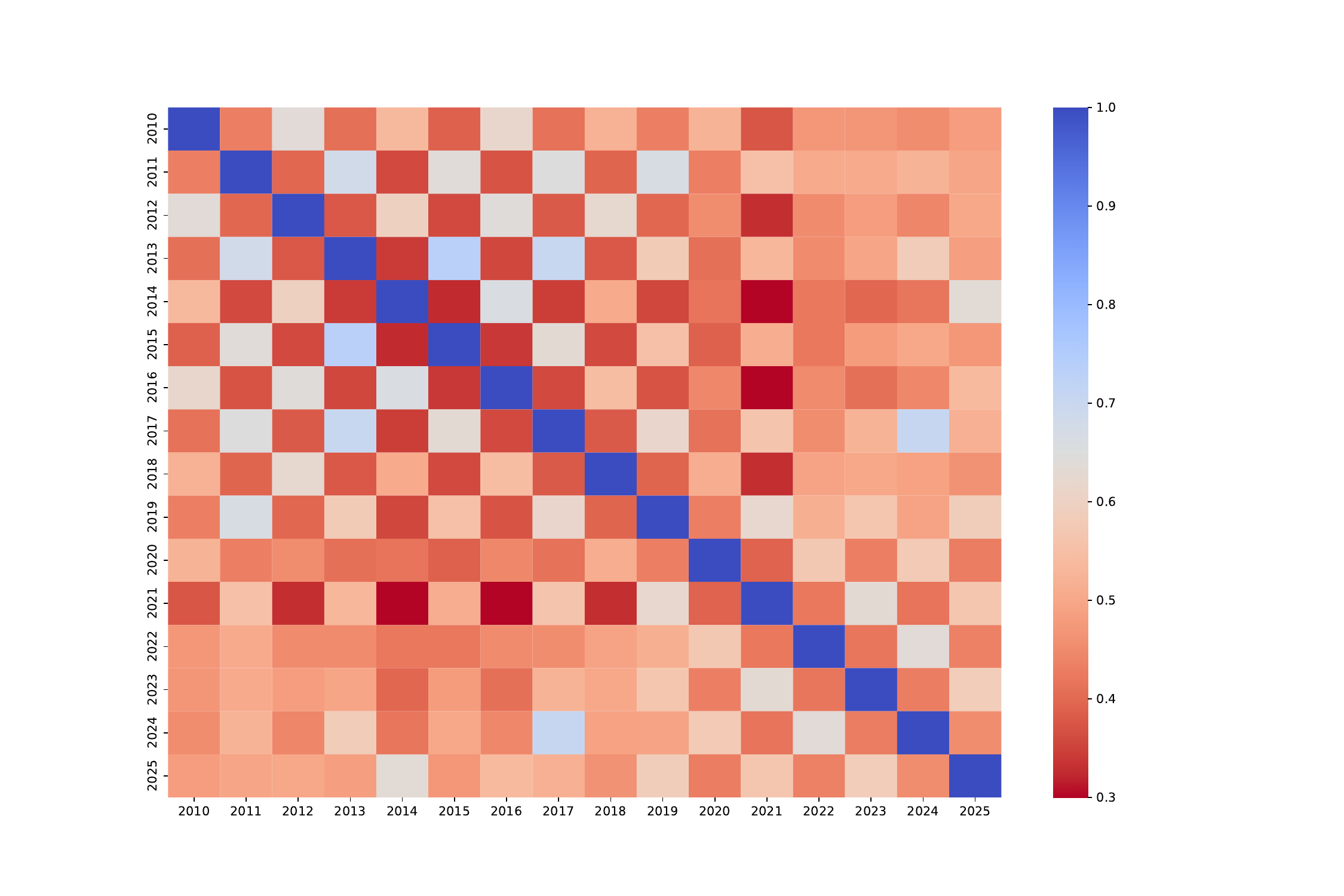}
  }
  \subfloat[Ticktack temporal aligned]
  {
      \includegraphics[height=4.3cm,width=5.4cm, trim=2.0cm 0.8cm 3.cm 1.0cm, clip]{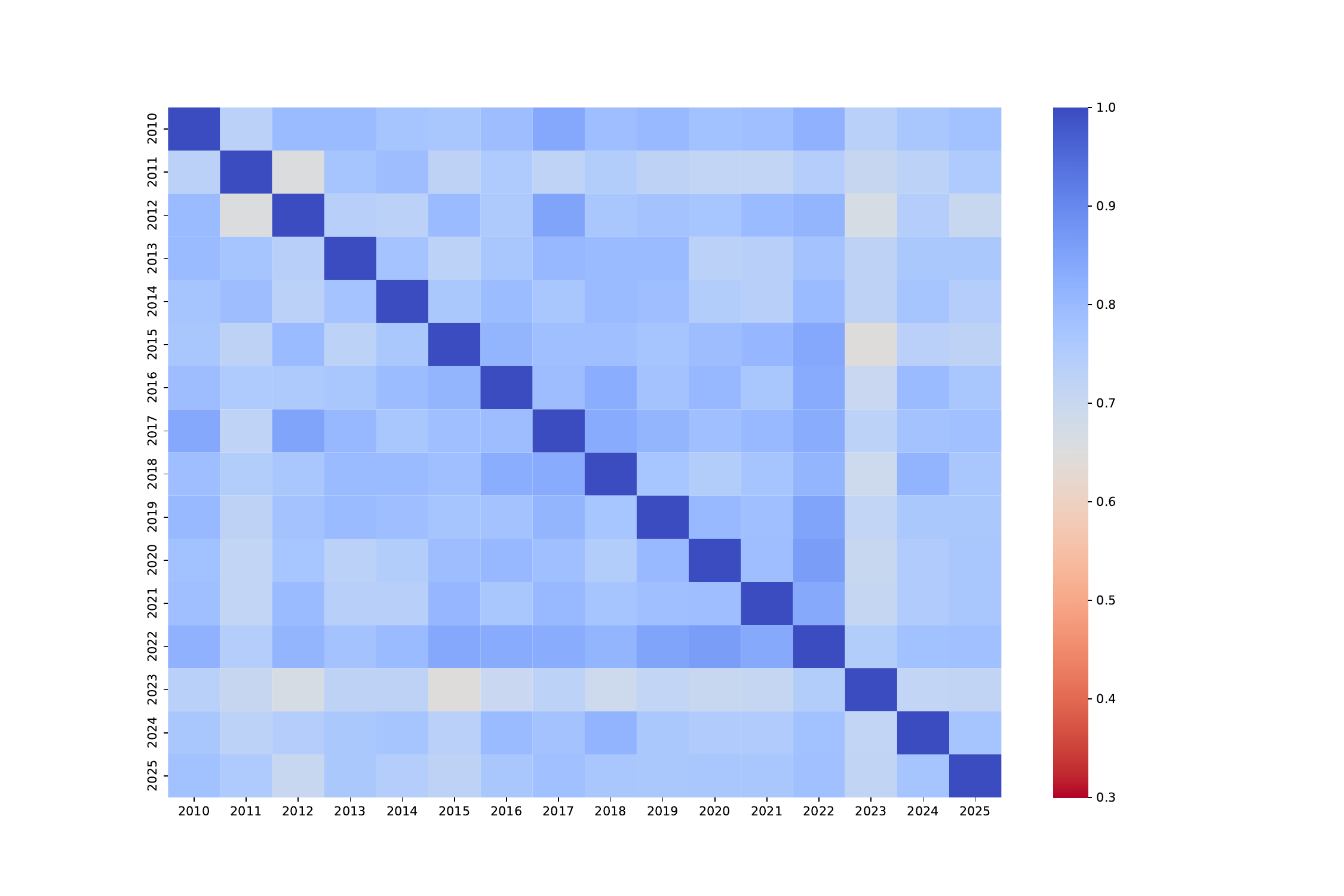}
  }
  \caption{Similarity between different years' representations, where years are chosen from 2010 to 2025. As the color of the cell approaches orange, it indicates a lower similarity between these two years' representations. (a) The embeddings of Gregorian years are generated by the base Qwen 3B. (b)  Sexagenary year time encoding $TE_(x_{i})$ and $TE_(y_{i})$ proposed in Eq. \eqref{eq_TE}, without post training. (c) Using the Ticktack post-trained Qwen-3b w/Ticktack to generate the representations of Gregorian years. }   
  \label{img_heatmap}          
\end{figure*}

\textbf{Performance on low resource years.} To analyze temporal reasoning ability in low-resource years, we separated the TempLS dataset into 500-year intervals, in addition to BCE and after 2000 periods, as shown in Figure \ref{img_dataset}. We hypothesize that the years of the BCE period have a comparatively low frequency of occurrence due to the long duration of seventy thousand five hundred years. As illustrated in Figure \ref{img_radar}, Ticktack achieves more significant enhancement in the low-resource years of BCE, intuitively displayed through the yellow line.


\textbf{Visualization of sexagenary year representations.} We use T-SNE \citep{t-sne} to visualize the multi-dimensional embeddings of Qwen-3B's outputs before and after temporal alignment with Ticktack, as illustrated in Figure \ref{img_tsne}. We sample 10,000 sentence vectors across ten-year periods from TempLAMA
("Geng Yin to Ji Hai").  It is distinctly evident that Ticktack-enhanced embeddings in Figure \ref{img_tsne}(b) exhibit clustering characteristics according to years. The pre-trained embeddings in Figure \ref{img_tsne}(a), on the other hand, are dispersed throughout the vector space, which may make the model more susceptible to temporal reasoning errors.

\begin{table}[ht]
\centering
\small
\caption{\label{table_abl} Zero-shot and few-shot results of LLMs measured on TempLS, by adding temporal encoding module.}
\setlength{\tabcolsep}{1mm}{
\begin{tabular}{c|r|ll}
\toprule[0.5mm]
\multicolumn{2}{c}{\multirow{2}{*}{\textbf{Tasks}}}  & \multicolumn{2}{c}{\textbf{TempLS}} \\ \cmidrule(r){3-3} \cmidrule(r){4-4}
\multicolumn{1}{l}{}   & \multicolumn{1}{l}{}               & \multicolumn{1}{l}{\textbf{Zero-shot}}                             & \multicolumn{1}{l}{\textbf{Few-shot}} \\ \hline
\multicolumn{2}{c}{\textbf{Model}} & \multicolumn{1}{c}{\textbf{Acc.}} & \multicolumn{1}{c}{\textbf{Acc.}} \\ \hline
\multirow{3}{*}{\textbf{\small{Qwen2.5-3B}}} & \textbf{Base}  & \multicolumn{1}{l}{62.37}   & \multicolumn{1}{l}{67.81}  \\
& \textbf{w/ encoding}  & \multicolumn{1}{l}{66.55\scriptsize{(+4.18)}}   & \multicolumn{1}{l}{66.13\scriptsize{(-1.68)}}  \\ 
\hline
\multirow{3}{*}{\textbf{\small{Qwen2.5-7B}}} & \textbf{Base}  & \multicolumn{1}{l}{73.66}   & \multicolumn{1}{l}{72.89}  \\
& \textbf{w/ encoding}  & \multicolumn{1}{l}{73.89\scriptsize{(+0.23)}}   & \multicolumn{1}{l}{79.94\scriptsize{(+7.05)}}   \\ 
\hline
\multirow{3}{*}{\textbf{\small{LLaMA2-7B}}} & \textbf{Base}  & \multicolumn{1}{l}{21.54}   & \multicolumn{1}{l}{48.52} \\
& \textbf{w/ encoding}  & \multicolumn{1}{l}{30.58\scriptsize{(+9.04)}}   & \multicolumn{1}{l}{50.98\scriptsize{(+2.46)}}  \\ 
\hline
\multirow{3}{*}{\textbf{\small{LLaMA2-13B}}} & \textbf{Base}  & \multicolumn{1}{l}{32.63}   & \multicolumn{1}{l}{20.88}  \\
& \textbf{w/ encoding}  & \multicolumn{1}{l}{30.68\scriptsize{(-1.95)}}   & \multicolumn{1}{l}{35.31\scriptsize{(+14.43)}}   \\
\bottomrule[0.5mm]
\end{tabular}}
\end{table}

\subsection{Ablation study}
\textbf{Similarity between different years representations.}  
Figure \ref{img_heatmap} depicts the distinguishability of the representation of years (2010-2025), generated by LLMs. Figure \ref{img_heatmap}(a) shows the representation similarity of Gregorian years from the Qwen 3B base, which has similar characteristics, making it difficult for the model to distinguish. Alternatively, our proposed sexagenary year expression with polar coordinate time encoding introduces enriched temporal information into the LLM, resulting in a clear distinction as depicted in Figure \ref{img_heatmap} (b). After post-training with Ticktack, the temporally aligned LLM improves its sensitivity to temporal information, as seen in Figure \ref{img_heatmap} (c). In comparison to the original base model, time embeddings become more discriminable, making LLMs more likely to recognize, contributing to increased performance on time-sensitive tasks.

\textbf{The impact of temporal encoding module.} To investigate the effect of the sexagenary year expression and encoding, we post train the LLMs by adopting temporal encoding of the polar coordinate represented sexagenary year (w/ encoding) with predict next token target that does not use temporal alignment. As shown in Table \ref{table_abl}, only the sexagenary year's temporal encoding facilitates the enhancement of time-sensitive reasoning capabilities.

\section{Conclusion}
In this study, we focus on addressing the temporal misalignment issues that often affect LLMs when dealing with long-span temporal information. We first introduce the sexagenary-cycle time expression leveraging the polar coordinate to provide a more uniform and consistent temporal embedding expression. Furthermore,
a temporal alignment method is proposed to enhance the LLMs’ alignment of learned knowledge to the related time period. Experimental results have validated the effectiveness of our method, demonstrating its ability to enhance the performance of LLMs in handling time-related tasks with long temporal spans.

\section{Limitations}
While we use the sexagenary cycle time expression to align long-term temporal data, we only consider the year granularity. We will explore using the sexagenary cycle to represent many granularities of time, such as month and day, in the future. In addition, since there are few long-span time-related benchmarks, we developed the TempLS dataset, which we collected from Baidu Baike, to measure LLMs' understanding of temporal information. The relatively small number of samples may limit the generalizability and robustness of the evaluation results. It would be interesting to develop novel benchmarks derived from a range of sources to increase the comprehensiveness and reliability of temporal information understanding assessments.

\bibliography{custom}

\appendix

\section{Appendix}
\label{sec:appendix}

\subsection{Datasets}
\label{datasets}
\textbf{TempLAMA} \citep{templama}: a time-sensitive question-answering dataset constructed based on the Wikidata temporal KB, is proposed to evaluate the model’s performance for time-dependent questions from 2010 to 2020. The generated answers of LLMs are evaluated by token-level micro-F1 and ROUGE-1 scores. 

\textbf{TempUN} \citep{ beniwal2024remember}: a large temporal QA dataset constructed by curating temporal information from "Our World in Data" website. TempUN is used to explore the model's ability to grasp factual knowledge, containing data for global issues like poverty, disease, hunger, climate change, war, existential risks, and inequality from 10,000 BCE to 2100 AD. The format of this task is a multiple choice question answering, with accuracy serving as the evaluation metric.

Samples of the three datasets are illustrated in Table \ref{tab:examples}.

\begin{CJK*}{UTF8}{gbsn}
\begin{table*}[bt]
\centering
\caption{Samples from TempLS, TempLAMA and TempUN.}
\label{tab:examples}
\setlength{\tabcolsep}{1.7mm}{
\begin{tabular}{c|rl}
\toprule[0.5mm]
\textbf{Dataset}  & \multicolumn{2}{c}{\textbf{Sample}} \\ 
\midrule
\multirow{3}{*}{\textbf{\small{TempLS}}} & \multirow{2}{*}{Question:}  & \multicolumn{1}{l}{\makecell[tl]{In 606 AD, \_\_\_\_sent an ambassador to pay tribute. \\  A: Khmer Emperor, B: Mikado, C: Goryeo Emperor, D: Vietnam Emperor}}  \\
& Answer: & \multicolumn{1}{l}{A} \\
\midrule
\multirow{2}{*}{\textbf{\small{TempLAMA}}} & Question:  & \multicolumn{1}{l}{In 2017, Alexander Hamilton is owned by \_X\_.}  \\
& Answer:  & \multicolumn{1}{l}{Crystal Bridges Museum of American Art} \\
\midrule
\multirow{4}{*}{\textbf{\small{TempUN}}} & \multirow{3}{*}{Question:}  & \multicolumn{1}{l}{\makecell[tl]{Which option is correct for the question:\\  In 2022, Private Civil Liberties Index in Iran was: \\ Options: A: 0.49, B: 0.38, C: 0.63, D: 0.34}}  \\
& Answer: & \multicolumn{1}{l}{D} \\
\bottomrule[0.5mm]
\end{tabular}}
\end{table*}
\end{CJK*}

\subsection{Visualization of sexagenary cycle time expression}
We also utilize T-SNE to project the multi-dimensional embeddings of LLaMA2-13B and Deepseek-7B’s outputs before and after temporal adaptation with Ticktack into a two-dimensional space for visualization. 

\begin{figure}[ht]
\centering
\includegraphics[height=8.5cm,width=5.8cm]{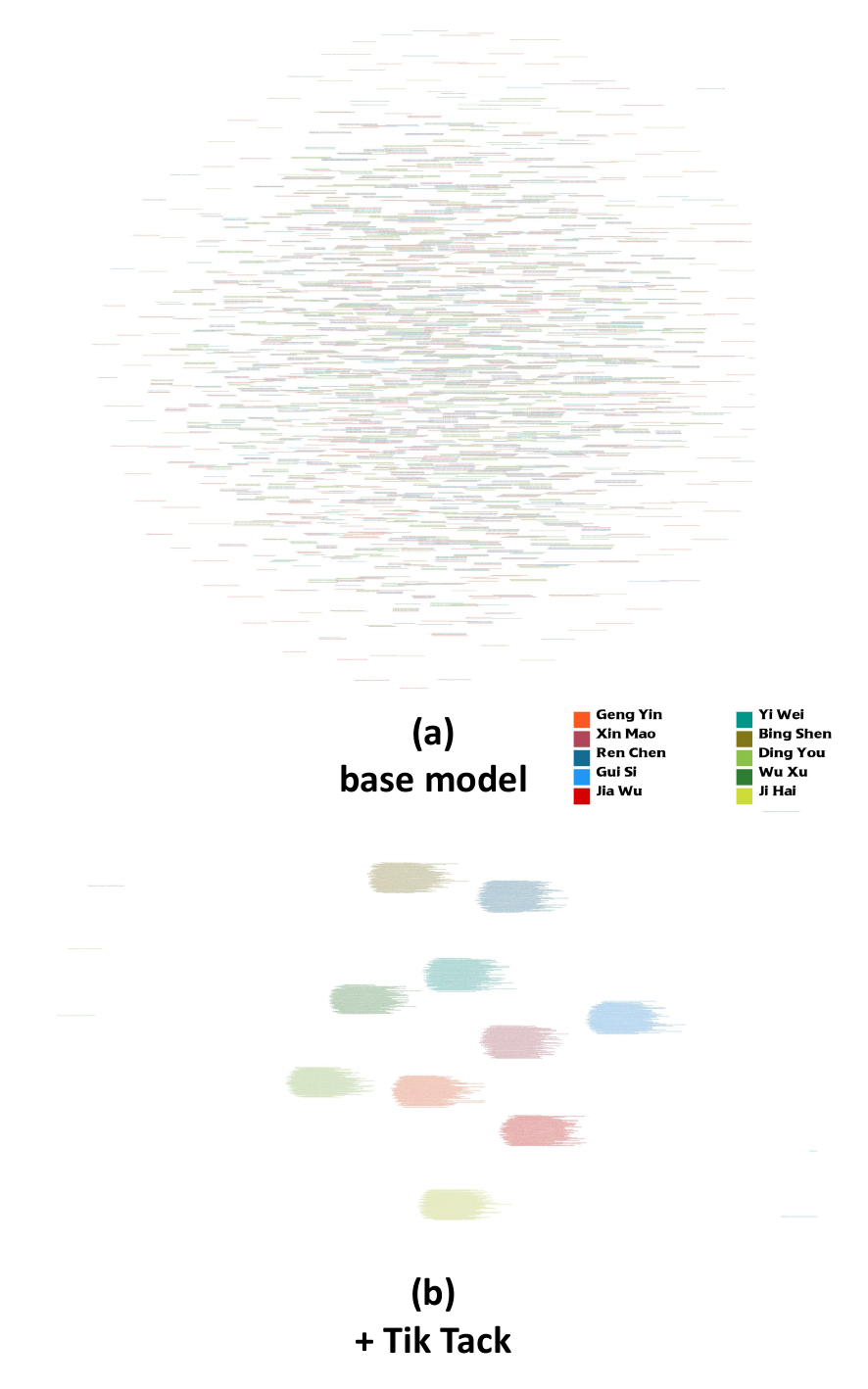}
\caption{The distribution of LLaMA2-13B's outputs of 10,000 sentence vectors, before and after post training by Ticktack. }
\label{img_tsne_llama}
\end{figure}

\begin{figure}[ht]
\centering
\includegraphics[height=9cm,width=5.7cm]{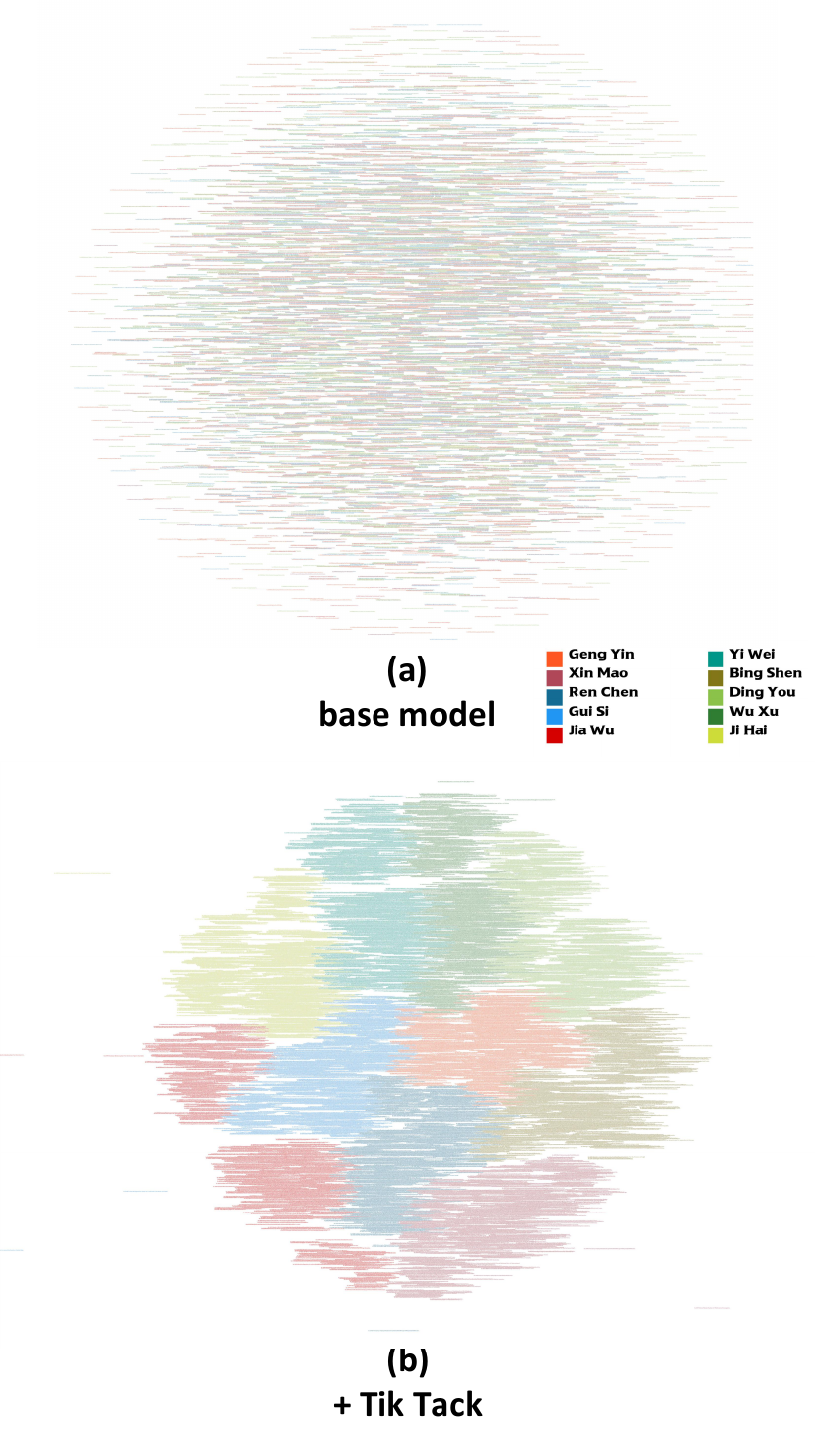}
\caption{The distribution of Deepseek-7B's outputs of 10,000 sentence vectors, before and after post training by Ticktack. }
\label{img_tsne_deepseek}
\end{figure}

As shown in Figure \ref{img_tsne_llama} and \ref{img_tsne_deepseek}, the expressions of all models also exhibit characteristics of aggregation based on the sexagenary cycle after post-training used Ticktack. However, compared to  Qwen-3B displayed in Figure \ref{img_tsne}, the intra-class distance in LLaMA2-13B is more concentrated, while the distance between inter-classes is more dispersed. This may be due to the different training data adopted by different LLMs.

\end{document}